\documentclass[lettersize,journal]{IEEEtran}
\usepackage{amsmath,amsfonts}
\usepackage{algorithmic}
\usepackage{algorithm}
\usepackage{array}
\usepackage[caption=false,font=normalsize,labelfont=sf,textfont=sf]{subfig}
\usepackage{textcomp}
\usepackage{stfloats}
\usepackage{url}
\usepackage{verbatim}
\usepackage{graphicx}
\usepackage{cite}

\usepackage{xcolor}
\usepackage{booktabs}
\usepackage{multirow}
\usepackage{makecell}
\usepackage{hyperref}

\begin{document}

\title{In-Context Translation: Towards Unifying Image Recognition, Processing, and Generation}

\author{Han Xue$^{\ast}$, Qianru Sun, Li Song,~\IEEEmembership{Senior Member,~IEEE,} Wenjun Zhang,~\IEEEmembership{Fellow,~IEEE,} and Zhiwu Huang
\thanks{Han Xue and Wenjun Zhang are with the Institute of Image Communication
and Network Engineering, Shanghai Jiao Tong University, Shanghai 200240,
China (e-mail: \{xue\_han, zhangwenjun\}@sjtu.edu.cn).}
\thanks{Qianru Sun is with the School of Information Systems, Singapore Management University, Singapore 178902 (e-mail: qianrusun@smu.edu.sg).}
\thanks{Li Song (Corresponding author) is with the Institute of Image Communication and Network Engineering, Shanghai Jiao Tong University, Shanghai 200240, China, and also with the MoE Key Laboratory of Artificial Intelligence, AI Institute, Shanghai Jiao Tong University, Shanghai 200240, China (e-mail: song\_li@sjtu.edu.cn).}
\thanks{Zhiwu Huang is with the School of Electronics and Computer Science, University of Southampton, Southampton, United Kingdom (e-mail: zhiwu.huang@soton.ac.uk).}
\thanks{$^{\ast}$ Part of this work was done when Han Xue served as a visiting Ph.D. student at Singapore Management University.}}

\markboth{Journal of \LaTeX\ Class Files,~Vol.~14, No.~8, August~2021}%
{Shell \MakeLowercase{\textit{et al.}}: A Sample Article Using IEEEtran.cls for IEEE Journals}


\maketitle

\begin{abstract}
We propose In-Context Translation (\texttt{ICT}), a general learning framework to unify visual recognition (\textit{e.g.}, semantic segmentation), low-level image processing (\textit{e.g.}, denoising), and conditional image generation (\textit{e.g.}, edge-to-image synthesis). 
Thanks to unification, \texttt{ICT} significantly reduces the inherent inductive bias that comes with designing models for specific tasks, and it maximizes mutual enhancement across similar tasks.
However, the unification across a large number of tasks is non-trivial due to various data formats and training pipelines. 
To this end, \texttt{ICT} introduces two designs. Firstly, it standardizes input-output data of different tasks into RGB image pairs, \textit{e.g.}, semantic segmentation data pairs an RGB image with its segmentation mask in the same RGB format. 
This turns different tasks into a general translation task between two RGB images.
Secondly, it standardizes the training of different tasks into a general in-context learning, where ``in-context'' means the input comprises an example input-output pair of the target task and a query image.
The learning objective is to generate the ``missing'' data paired with the query. The implicit translation process is thus between the query and the generated image.
In experiments, \texttt{ICT} unifies ten vision tasks and showcases impressive performance on their respective benchmarks. Notably, \texttt{ICT} performs well across three major categories of computer vision tasks, while its two competitors (Painter and PromptDiffusion) are only effective in at most two of these task categories. In addition, compared to its competitors, \texttt{ICT} trained on only 4 RTX 3090 GPUs is shown to be more efficient and less costly in training.
\end{abstract}


\section{Introduction}
\label{sec:intro}

The natural language processing (NLP) community has witnessed the great success of large language models (LLMs)~\cite{DBLP:conf/naacl/DevlinCLT19,radford2019language,raffel2020exploring,brown2020language,chowdhery2023palm}.
A compelling feature is that LLMs can serve as a generalist to handle a wide range of downstream tasks within a single framework~\cite{brown2020language}.
This can be attributed to 1)~emerging abilities of large-scale training~\cite{wei2022emergent}, 2)~a unified task formulation, \textit{e.g.}, a variety of NLP tasks can be consistently framed as text completion~\cite{brown2020language}, and 3)~in-context learning that standardizes the adaptation process from a pre-trained model to downstream tasks~\cite{brown2020language,liu2021makes,min2022rethinking,rubin2021learning,wei2021finetuned,alayrac2022flamingo}.

In the computer vision (CV) community, a unified framework for different tasks is also a long-standing aspiration. The unification is appealing because it side-steps task-specific designs, minimizes the inductive bias inherent in crafting models for different tasks, and maximizes the potential of mutual enhancement among related tasks.
However, the progress of such unification in CV lags much behind NLP, due to two main \textbf{C}hallenges.
\textbf{C1}: NLP has the language token as an intuitive and unified data format for all tasks. However, CV encompasses highly heterogeneous data formats, \textit{e.g.}, images, semantic segmentation masks, keypoints, depth maps, \textit{etc}.
Such variation impedes the unification of expert models of different CV tasks.
\textbf{C2}: Most NLP tasks can be 
easily converted into a sequence completion task~\cite{brown2020language} as language itself is sequential.
In CV, the popular solution for image completion is Masked Autoencoders (MAE)~\cite{he2022masked}. The objective of MAE is to acquire an image representation by reconstructing the original image from a version
where a random subset of image patches is masked out.
This representation can be tailored to various specific discrimination tasks through fine-tuning. However, it is not intuitive how it can be applied to unify other categories of vision tasks, such as conditional image generation.

\begin{figure*}[!t]
  \centering
    \includegraphics[width=1.0\linewidth]{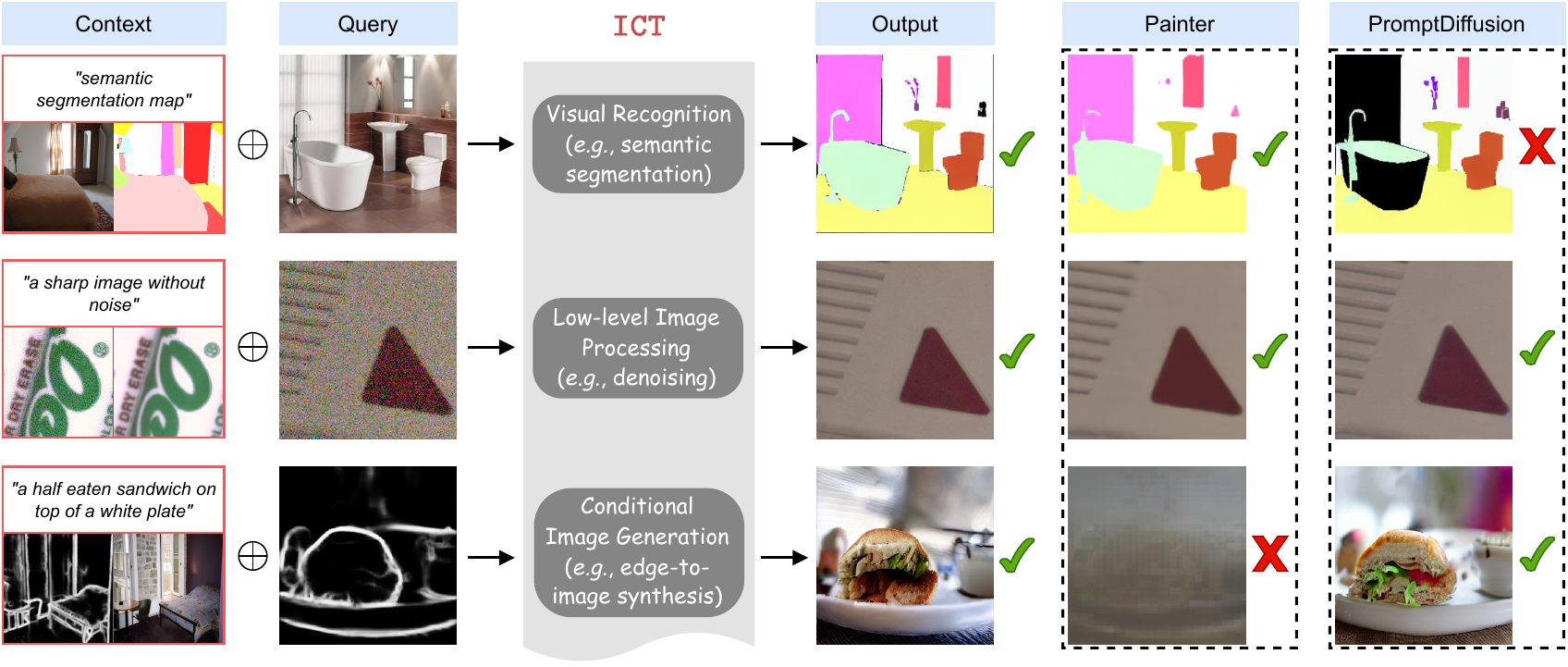}
  \caption{\texttt{ICT} shows impressive results in unifying three distinct categories of vision tasks within a single framework.
  Each task is implicitly instructed by a random input-output pair (with an optional text prompt). 
  Compared to its competitors such as Painter~\cite{wang2023images} and PromptDiffusion~\cite{wang2023context}, \texttt{ICT} unifies more categories of tasks. Please note that the inputs of competitors are different, and see more detailed comparisons in experiments.}
  \label{fig:teaser}
\end{figure*}

We tackle these challenges by introducing a new learning framework called In-Context Translation (\texttt{ICT}).
\texttt{ICT} unifies three distinct categories of vision tasks---visual recognition (\textit{e.g.}, semantic segmentation), low-level image processing (\textit{e.g.}, denoising), and conditional image generation (\textit{e.g.}, edge-to-image synthesis)---onto an in-context image translation framework.
Specifically, to solve \textbf{C1}, \texttt{ICT} formulates the input and output of different tasks all as RGB images. For instance, for the depth estimation task, \texttt{ICT} converts the one-channel depth map into an RGB image by linearly transforming the depth values to $[0, 255]$ and replicating the transformed map three times.
By this, \texttt{ICT} paves the way to turning different tasks into an \emph{image translation task} between two RGB images, making it possible to yield a unified model.
To solve \textbf{C2}, \texttt{ICT} builds an image translation framework integrated with the idea of in-context learning where the ``context'' defines the target task.
The input is a spatial concatenation of an input-output pair from a target task (\textit{i.e.}, context) and a query image. The learning objective is to translate this query into its paired data in the same task.

Thanks to \texttt{ICT}'s unifying designs, in practice, it can be deployed to train various models according to the degree of unification. 
1)~When trained on joint data of different categories of tasks (\textit{e.g.}, semantic segmentation, image denoising, and depth-to-image synthesis), it yields a compact model that can generate outputs for every trained task.
2)~When trained on joint data of a specific category (\textit{e.g.}, mask-to-image and depth-to-image data for the category of conditional image generation), it derives a single-category model.
3)~When trained on data of an individual task, it learns a dedicated single-task model. This is a special case where the model itself is not unified.
\emph{We highlight that the above three models are trained in different regimes and datasets, yielding distinct model parameters, but the training framework (including input-output format, backbone, training method, etc.) always remains the same, underscoring the ``unifying'' essence}.

The implementation of \texttt{ICT} follows 
two successful \textbf{P}ractices of NLP generalists: \textbf{P1} a large-scale pre-trained backbone, and \textbf{P2} context-guided fine-tuning.
For \textbf{P1}, we build \texttt{ICT} on Stable Diffusion (SD)~\cite{rombach2022high} given that 1)~SD is a generative model to yield high-quality image output, which cannot be achieved by pre-trained ViT~\cite{dosovitskiy2020image} or MAE~\cite{he2022masked}, and 2)~SD is trained on billion-level web data LAION-5B~\cite{schuhmann2022laion} and can provide a comprehensive prior that integrates a deep understanding of both visual and linguistic signals.
For \textbf{P2}, we convert the conventional datasets of ten vision tasks into context-based datasets and fine-tune the U-Net of SD on them using the proposed \texttt{ICT} framework. 
Thanks to the linguistic ability of SD, we can optionally add a textual prompt (either task-level or instance-level) during fine-tuning, providing flexible control over the results.

Our experiments showcase extensive results of \texttt{ICT} in three types consisting of ten tasks in total.
Our key observation is that \texttt{ICT} exhibits promising performance on all categories of tasks, while its competitors Painter~\cite{wang2023images} and PromptDiffusion~\cite{wang2023context} collapse on at least one of them (Fig. \ref{fig:teaser}).
Our contribution is thus three-fold:
\begin{itemize}
\item{A general learning framework that can cope with visual recognition, low-level image processing, and conditional image generation tasks.}
\item{A new in-context translation method that can be built on Stable Diffusion (SD), adapting SD's pre-trained knowledge for downstream tasks in a unified fashion.}
\item{Extensive experiments on ten vision tasks across three categories and for three types of model training, by which we hope to spur more interesting research on how to induce a profound understanding of vision tasks through a unified learning framework.}
\end{itemize}

\section{Related Works}
\label{sec:rel}

\subsection{Unified Vision Models}
\label{sec:rel_sub1}
Encouraged by the success of language generalist models~\cite{brown2020language,chowdhery2023palm,touvron2023llama}, training a generalist model to unify different computer vision tasks has attracted significant interest in recent years.
We identify three key components of designing a unified vision model: 1)~data format, 2)~backbone, and 3)~training pipeline. Regarding these, we discuss related works and their primary differences from ours.

Some attempts~\cite{wang2022ofa,chen2021pix2seq,kolesnikov2022uvim,chen2022unified,lu2022unified,lu2024unified,bai2024sequential} map the input image into discrete representations and implement sequence-to-sequence learning in the discrete space to handle different tasks.
Two representative works, Unified IO~\cite{lu2022unified} and Unified-IO 2~\cite{lu2024unified}, homogenize various vision data modalities into a sequence of discrete vocabulary tokens and utilizes VQGAN~\cite{esser2021taming} to support dense prediction tasks. This unification of data format via the discretization process leads to lossy data compression, making the solution suboptimal. In contrast, our unified RGB image format better suits vision tasks.

Uni-Perceiver series~\cite{zhu2022uni,zhu2022uni-moe,li2023uni} introduce a unified maximum likelihood estimation pipeline for different tasks with various modalities but they have not been proven successful for image generation tasks. Instead of training from scratch, we leverage a pre-trained generative model as the backbone, and demonstrate strong performance, especially for conditional image generation.

Some recent works~\cite{liu2023llava,wang2023gpt4video,DBLP:conf/icml/Wu0Q0C24,wu2024visionllm} exploit the LLM-centric solution: they align different modalities into a unified space defined by a pre-trained LLM (\textit{e.g.}, Vicuna~\cite{zheng2024judging}) and invoke expert models to execute different tasks (\textit{e.g.}, Stable Diffusion~\cite{rombach2022high} for text-to-image generation and SEEM~\cite{zou2024segment} for image segmentation). However, these vision LLMs still rely on various task-specific models, which is highly inefficient when the number of tasks increases. In addition, sophisticated instruction-based data curation is needed for training.

Similar to our \texttt{ICT}, a bunch of  works~\cite{bar2022visual,wang2023images,wang2023context,DBLP:conf/iclr/GanPSPA24,geng2024instructdiffusion,fan2024toward} utilize the RGB image as a general data format to unify vision tasks.
MAE-VQGAN~\cite{bar2022visual} and Painter~\cite{wang2023images} propose a masked image generation framework, where the input image and an example pair are stitched together and the model needs to predict the masked region (similar to MAE~\cite{he2022masked}).
They adopt a pre-trained ViT-based MAE as the backbone, limiting their performance in image generation. \texttt{ICT} does not suffer from this issue due to the generative backbone we use.
In addition, they apply random masking of image patches during training while \texttt{ICT} is trained to predict the \emph{whole} desired output image that is more in line with the inference scenario.
\texttt{ICT} supports the usage of textual prompt to enable more controlled image generation, which is beyond the reach of MAE-VQGAN and Painter.

PromptDiffusion~\cite{wang2023context} incorporates in-context learning into a pre-trained diffusion model. It sums up the features of context and query images to perform in-context modeling. We argue that this pixel-wise sum operation is ineffective for in-context learning, since this makes it challenging for the model to reconstruct or comprehend the context from the summed values during both training and inference. Additionally, a study by~\cite{chen2023improving} 
notes that PromptDiffusion struggles with image in-context tasks when lacking a textual prompt. In contrast, our \texttt{ICT} avoids sum operations and instead, spatially concatenates context and query images. It gets in-context instructions directly from the original images.

InstructCV~\cite{DBLP:conf/iclr/GanPSPA24} and InstructDiffusion~\cite{geng2024instructdiffusion} learn a unified model based on a generative backbone and cast different vision tasks as text-guided image editing. Nevertheless, they heavily rely on delicate prompt design, limiting their applications to vision tasks that are difficult to instruct by using human language (\textit{e.g.}, InstructDiffusion cannot handle the task of depth estimation). \texttt{ICT} refrains from this issue by applying a simple image-based context to inform the desired task.

\subsection{Diffusion Models}
\label{sec:rel_sub2}
Diffusion models~\cite{sohl2015deep,ho2020denoising,song2020denoising,song2020score,song2020improved} have recently become the primary choices for generative modeling of data. Following the definition in Denoising Diffusion Probabilistic Models (DDPM)~\cite{ho2020denoising}, a diffusion model consists of a forward diffusion process that gradually adds noise to data and a reverse denoising process that reconstructs desired data from the noise.
Recent methods based on diffusion models achieve state-of-the-art results in many vision tasks~\cite{croitoru2023diffusion}, including image/video generation~\cite{dhariwal2021diffusion,luo2023videofusion}, image editing~\cite{meng2021sdedit,brooks2023instructpix2pix}, image-to-image translation~\cite{xia2024diffusion}, low-level image processing~\cite{saharia2022image,wang2022zero}, \textit{etc}.
Notably, large-scale text-to-image diffusion models show compelling ability~\cite{nichol2021glide,saharia2022photorealistic,rombach2022high,balaji2022ediffi} to generate high-fidelity visual content. These pre-trained models are broadly utilized in applications such as concept customization~\cite{gal2022image,ruiz2023dreambooth,sun2024create} and image composition~\cite{lu2023tf}. Some recent works~\cite{li2023your,clark2024text,zhao2023unleashing,xu2023open,tian2023diffuse} further reveal the potential of applying pre-trained diffusion models to discriminative tasks, which encourages us to build a general framework that unifies generative and discriminative model training.

\subsection{In-Context Learning}
\label{sec:rel_sub3}
GPT-3~\cite{brown2020language} has demonstrated the ability of pre-trained LLMs to adapt to various downstream NLP tasks given example pairs. After that, there have been efforts in exploring the ability of in-context learning~\cite{liu2021makes,min2022rethinking,rubin2021learning,mishra2021cross,wei2021finetuned,sanh2021multitask}.
A prominent track of works~\cite{wang2022self,ouyang2022training,chung2024scaling,wu2023improving} fine-tunes language models on a collection of datasets described via instructions (\textit{e.g.}, task prompt, demonstration examples, and constraints), and attains significant improvement of the model's generalization on unseen tasks.
This in-context learning pipeline has recently been introduced to vision-language tasks as well~\cite{alayrac2022flamingo,liu2024visual,gao2023llama,li2023otter}. A representative work, Flamingo~\cite{alayrac2022flamingo}, bridges pre-trained vision and language models by fine-tuning them on text-image instruction-following data and showcases impressive few-shot results in a variety of tasks such as image captioning and visual question-answering.
By comparison, \texttt{ICT} exploits a new approach of in-context translation, which is based on the use of an image-label pair as well as an optional prompt for both image- and text-level context.
\section{In-Context Translation (\texttt{ICT})}
\label{sec:approach}

We propose In-Context Translation (\texttt{ICT}), a unified framework that copes with visual recognition, low-level image processing, and conditional image generation tasks.
It is to learn a mapping function $f$ that conducts the same in-context image translation process for different tasks:
\begin{equation}
  f(E_{in}, E_{out}, y, I_{query})=I_{gt},
  \label{eq1}
\end{equation}
where ($E_{in}$, $E_{out}$) represents an example pair of input and output data (called ``context'' in this paper) sampled from the target task.
Taking semantic segmentation as an example, $E_{in}$ and $E_{out}$ represent an RGB image and its corresponding segmentation map in RGB format, respectively. 
$y$ is a textual prompt to represent the task and/or to provide instance-level information (which is optional in the implementation of \texttt{ICT}). $I_{query}$ is the query image,
and $I_{gt}$ is the corresponding ground truth image to be translated to in the target task.

To learn function $f$, we first construct context-based training data by converting the input-output data into RGB images consistently across different vision tasks (Section~\ref{sec:approach_sub1}). 
Then, we build the \texttt{ICT} framework on top of SD~\cite{rombach2022high} and offer different training strategies according to the unification degrees (Section~\ref{sec:approach_sub2}). For example, for the most unification, each model training iteration uses a batch of context-based data sampled from ten vision tasks (across three types).

\begin{figure*}[!t]
  \centering
    \includegraphics[width=0.86\linewidth]{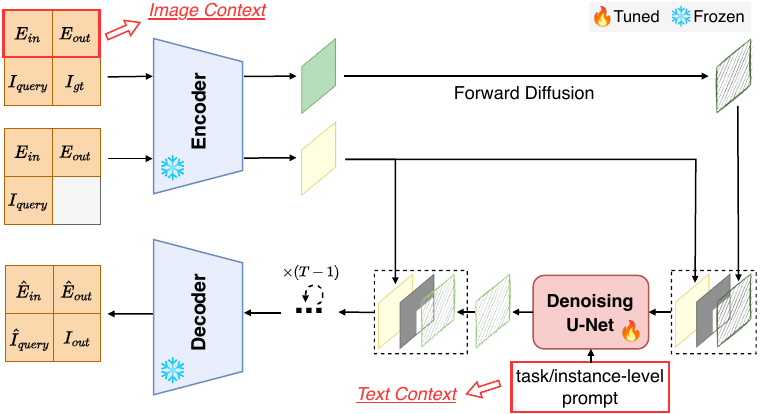}
  \caption{An overview of the proposed framework. It utilizes a pre-trained SD to perform in-context image translation. The ground truth is established as a grid image where each row is an input-output pair from the same task. The first row composed by $E_{in}$ and $E_{out}$ serves as the image context, and the model is trained to predict $I_{gt}$ paired to the query image $I_{query}$. At inference time, we crop out the lower right region of the infilled output as the final result $I_{out}$.}
  \label{fig:framework}
\end{figure*}

\subsection{Context-based Data Construction}
\label{sec:approach_sub1}

\texttt{ICT} aims for the unification of
three types of tasks: visual recognition, low-level image processing, and conditional image generation. In the following, we introduce the specific tasks we focus on and elaborate on the process of constructing context-based data utilizing conventional datasets. 
The key idea is converting all data consistently into RGB images to streamline the spatial alignment of context and query images for training ICT models.

\noindent \textbf{Visual Recognition.}
We consider three representative recognition tasks, including semantic segmentation, depth estimation, and keypoint detection.
1)~Semantic segmentation is a dense classification task wherein the input is an image and the output is per-pixel semantic labels. We transfer the output to an RGB image by assigning a color to each pixel according to a predefined semantic-color codebook, ensuring that each semantic class has its unique RGB value~\cite{wang2023images}. During inference, we obtain the predicted semantic class of each pixel by finding its ``nearest'' color in the codebook. 
2)~Depth estimation is a dense regression task that needs to predict the depth value of each pixel of the input image. To convert the output, \textit{i.e.}, one-channel depth map, into an RGB image, we linearly scale the range of pixel values to $\left[0, 255\right]$ and replicate it three times to yield a three-channel ``image''.
3)~Keypoint detection aims to locate key object components on the input image. We formulate the output as a RGB image by first drawing squares (each square occupies 9$\times$9 pixels) whose center is the location of keypoints, and then rendering each square with a unique color associated with the key component.

To exploit the pre-trained knowledge of SD in understanding linguistic signals, we use textual prompts as text context. This prompt could be either \textit{task-level} (\textit{e.g.}, ``a semantic segmentation map'') or \textit{instance-level} (\textit{e.g.}, ``a semantic segmentation map of a man sitting on a swing in a room''). In practice, the latter is obtained through an off-the-shelf image captioning tool~\cite{li2022blip}.
Please note that we do not use the instance-level prompt (except for semantic segmentation) when comparing with competitors for fairness.

\noindent \textbf{Low-level Image Processing.}
We consider three typical tasks including image denoising, image deraining, and low-light image enhancement. The input and output of these tasks are RGB images, so we leave them unchanged when generating the context-based datasets.
Similar to visual recognition tasks, we use two kinds of textual prompt: 1)~\textit{task-level} (\textit{e.g.}, ``a sharp image without noise''), and 2)~\textit{instance-level} (\textit{e.g.}, ``a sharp image of a bathroom with a toilet, sink, and bathtub'').
Please note that we do not use the instance-level prompt when comparing with competitors for fairness.

\noindent \textbf{Conditional Image Generation.}
This type of task requires generating realistic images from conditions with sparse semantic labels, greatly differing from visual recognition and low-level image processing tasks. In this work, we consider four popular tasks, including mask-to-image, depth-to-image, pose-to-image, and edge-to-image synthesis. 
Inputs from the first three tasks can be converted to RGB format in the same way as used in visual recognition. For the edge-to-image synthesis task, we adopt the edge detection model provided by ControlNet~\cite{zhang2023adding} to generate HED edge maps~\cite{xie2015holistically} as the input.
The captions of output images (\textit{e.g.}, ``a cat sleeping on top of a pair of shoes'') are used as the text context.

\subsection{Training Framework}
\label{sec:approach_sub2}

We implement \texttt{ICT} to run an in-context image translation process on top of SD.
SD is a text-to-image model that incorporates a diffusion process in the latent space of a pre-trained autoencoder. 
Specifically, a denoising U-Net is trained to fit the distribution of latent codes, which models the reverse diffusion process. Taking the noisy latent and the time step as input, this U-Net is further conditioned on the textual embeddings extracted by the text encoder of CLIP~\cite{radford2021learning}.
Visual and textual embeddings interact by cross-attention inside the U-Net to produce the output of the current time step. 
During inference, SD performs iterative reverse diffusion on a randomly sampled noise to generate an image that faithfully adheres to the input text.
In our work, to fulfill (\ref{eq1}) for various visual tasks, we 
fine-tune the U-Net on our context-based datasets.

\begin{table*}[!t]
\setlength{\tabcolsep}{6.6pt}
\caption{Task and Dataset Configuration}
\label{tab:datasets}
  \centering
    \begin{tabular}{cccc}
    \toprule
    Task & Dataset & Training images & Testing images \\
    \midrule
    Semantic segmentation & ADE20K~\cite{zhou2017scene} & 20,210 & 2,000 \\
    Depth estimation & NYUv2~\cite{silberman2012indoor} & 24,231 & 654 \\
    Keypoint detection & COCO~\cite{lin2014microsoft} & 149,781 & 6,352 \\
    Image deraining & Merged deraining datasets~\cite{fu2017removing,yang2017deep,zhang2019image,zhang2018density,li2016rain} & 13,712 & 4,300 \\
    Image denoising & SIDD~\cite{abdelhamed2018high} & 96,000 & 1,280 \\
    Low-light image enhancement & LoL~\cite{wei2018deep} & 485 & 15 \\
    Mask-to-image synthesis & ADE20K~\cite{zhou2017scene} & 20,210 & 2,000 \\
    Depth-to-image synthesis & NYUv2~\cite{silberman2012indoor} & 24,231 & 654 \\
    Pose-to-image synthesis & COCO~\cite{lin2014microsoft} & 149,781 & 6,352 \\
    Edge-to-image synthesis & COCO~\cite{lin2014microsoft} & 118,287 & 5,000 \\
    \bottomrule
  \end{tabular}
\end{table*}

As shown in Fig.~\ref{fig:framework}, the image context (an example pair from a vision task, $E_{in}$ and $E_{out}$) is concatenated with another pair from the same task ($I_{query}$ and $I_{gt}$) to compose a grid image as the actual ground truth. During training, the input to the denoising U-Net comprises 3 components: 1)~the noisy latent embedding of ground truth, 2)~the latent embedding of a masked image $m$ similar to the ground truth but with $I_{gt}$ masked out, and 3)~the binary mask $b$ indicating the masked region.
The latter two serve as conditions to provide the model with context around the masked region and the location of the specific area to be infilled.
Text context is sent to the text encoder and the extracted textual embeddings are injected into the denoising U-Net via cross-attention.
With these contexts, the model is tuned to perform image completion, \textit{i.e.}, to generate the masked region. The training objective is the denoising loss of diffusion modeling: 
\begin{equation}
  \mathcal{L}(\theta)=\mathbb{E}_{z,m,b,y,\epsilon,t}\Bigl[\lVert\epsilon-\epsilon_{\theta}(z_{t},t,\mathcal{E}(m),b,c_{\phi}(y))\rVert_{2}^{2}\Bigr],
  \label{eq2}
\end{equation}
where $z$ is the latent code extracted from the ground truth, $y$ is the input text, $\epsilon\sim\mathcal{N}(0,1)$ is a noise term, $t$ is the time step, $\epsilon_{\theta}$ is the denoising U-Net, $z_{t}$ is the noisy version of $z$ at time $t$, $\mathcal{E}$ is the VAE encoder, and $c_{\phi}$ is the text encoder. We fine-tune the U-Net and keep the text encoder and the autoencoder frozen.

Note that previous inpainting-based models~\cite{bar2022visual,wang2023images} mask a portion of image patches---following the word masking method in NLP. However, we argue that such patch-level inpainting is inefficient for gaining a holistic and profound understanding of vision tasks, 
because the correlation between pixels is much stronger than that between words (\textit{e.g.}, the redundancy presented in images makes the model readily inpaint a patch according to its neighboring patches).
To tackle this, we mask the \textit{whole} output image and force the model to predict it during training, which is consistent with the inference scenario. We will show later that this new strategy fosters a better connection between visual features and semantics. This finding is in line with \cite{he2022masked} that masking a higher portion of random patches yields a more meaningful visual representation. This also implies the inherent difference between our generative modeling and the masked image modeling used by previous methods~\cite{bar2022visual,wang2023images}.
\section{Experiments}
\label{sec:exp}

\subsection{Experimental Settings}

\begin{table*}[!t]
\setlength{\tabcolsep}{3.0pt}
\caption{Comparison Results on Visual Recognition. $^{\ast}$: Specialized Methods for Each Task. $^{\ddagger}$: Officially Trained Painter Model Using 32$\times$ the Computing Power of \texttt{ICT}. $^{\dagger}$: Retrained Painter Using Its Official Code under the Same Computing Resources as \texttt{ICT}. $^{\S}$: A Concurrent Work (Results Are Obtained Using the Official Weights). \textbf{Bold}: Best. \underline{Underline}: Second Best. We Ignore Specialized Models When Ranking Best and Second Best and This Applies to All Tables. The Results of \texttt{ICT} Are Reported as the Average Scores and Standard Deviations across Three Trials
}
\label{tab:comp-prediction}
  \centering
    \begin{tabular}{ccccccc}
    \toprule 
    \multirow{2.5}*{Method} & ~ & Segmentation & ~ & \multicolumn{3}{c}{Depth estimation} \\
    \cmidrule{3-3}
    \cmidrule{5-7}
    ~ & ~ & mIoU$\uparrow$ & ~ & RMSE$\downarrow$ & REL$\downarrow$ & $\delta_{1}$$\uparrow$ \\
    \midrule
    \textcolor{gray}{OneFormer$^{\ast}$~\cite{jain2023oneformer}} & ~ & \textcolor{gray}{58.8} & ~ & \textcolor{gray}{-} & \textcolor{gray}{-} & \textcolor{gray}{-} \\
    \textcolor{gray}{Mask2Former$^{\ast}$~\cite{cheng2022masked}} & ~ & \textcolor{gray}{57.7} & ~ & \textcolor{gray}{-} & \textcolor{gray}{-} & \textcolor{gray}{-} \\
    \textcolor{gray}{ZoeDepth$^{\ast}$~\cite{bhat2023zoedepth}} & ~ & \textcolor{gray}{-} & ~ & \textcolor{gray}{0.270} & \textcolor{gray}{0.075} & \textcolor{gray}{0.955} \\
    \textcolor{gray}{BinsFormer$^{\ast}$~\cite{li2022binsformer}} & ~ & \textcolor{gray}{-} & ~ & \textcolor{gray}{0.330} & \textcolor{gray}{0.094} & \textcolor{gray}{0.925} \\
    \textcolor{gray}{Painter$^{\ddagger}$~\cite{wang2023images}} & ~ & \textcolor{gray}{49.9} & ~ & \textcolor{gray}{0.288} & \textcolor{gray}{0.080} & \textcolor{gray}{0.950} \\
    Painter$^{\dagger}$~\cite{wang2023images} & ~ & \underline{32.2} & ~ & \textbf{0.316} & \textbf{0.087} & \textbf{0.935} \\
    PromptDiffusion~\cite{wang2023context} & ~ & 18.2 & ~ & 0.746 & 0.171 & 0.799 \\
    InstructCV$^{\S}$~\cite{DBLP:conf/iclr/GanPSPA24} & ~ & - & ~ & 0.666 & 0.212 & 0.480 \\
    Unified-IO 2~\cite{lu2024unified} & ~ & - & ~ & 0.423 & - & - \\
    \texttt{ICT-st} & ~ & \textbf{33.4} $\pm$ 0.4 & ~ & \underline{0.420} $\pm$ 0.005 & \underline{0.135} $\pm$ 0.004 & \underline{0.857} $\pm$ 0.006 \\
    \bottomrule
  \end{tabular}
\end{table*}

\begin{table*}[!t]
\setlength{\tabcolsep}{2pt}
\caption{Comparison Results on Low-Level Image Processing. $^{\ast}$: Specialized Methods for Each Task. $^{\ddagger}$: Officially Trained Painter Model Using 32$\times$ the Computing Power of \texttt{ICT}. $^{\dagger}$: Retrained Painter Using Its Official Code under the Same Computing Resources as \texttt{ICT}. $^{\uplus}$: Following InstructDiffusion~\cite{geng2024instructdiffusion}, It Directly Reconstructs the Ground Truth via the Autoencoder of Pre-trained SD, and the Corresponding Results Indicate the Upper Bound of \texttt{ICT}. \textbf{Bold}: Best. \underline{Underline}: Second Best
}
\label{tab:comp-lowlevel}
  \scriptsize
  \centering
    \begin{tabular}{ccccccccccccc}
    \toprule 
    \multirow{2.5}*{Method} & ~ & \multicolumn{3}{c}{Deraining} & ~ & \multicolumn{3}{c}{Denoising} & ~ & \multicolumn{3}{c}{Enhancement} \\
    \cmidrule{3-5}
    \cmidrule{7-9}
    \cmidrule{11-13}
    ~ & ~ & PSNR$\uparrow$ & SSIM$\uparrow$ & LPIPS$\downarrow$ & ~ & PSNR$\uparrow$ & SSIM$\uparrow$ & LPIPS$\downarrow$ & ~ & PSNR$\uparrow$ & SSIM$\uparrow$ & LPIPS$\downarrow$ \\
    \midrule
    \textcolor{gray}{Restormer$^{\ast}$~\cite{zamir2022restormer}} & ~ & \textcolor{gray}{33.96} & \textcolor{gray}{0.935} & \textcolor{gray}{0.074} & ~ & \textcolor{gray}{40.02} & \textcolor{gray}{0.960} & \textcolor{gray}{0.198} & ~ & \textcolor{gray}{-} & \textcolor{gray}{-} & \textcolor{gray}{-} \\
    \textcolor{gray}{MIRNet-v2$^{\ast}$~\cite{zamir2022learning}} & ~ & \textcolor{gray}{-} & \textcolor{gray}{-} & \textcolor{gray}{-} & ~ & \textcolor{gray}{39.84} & \textcolor{gray}{0.959} & \textcolor{gray}{0.203} & ~ & \textcolor{gray}{24.74} & \textcolor{gray}{0.851} & \textcolor{gray}{0.116} \\
    \textcolor{gray}{Painter$^{\ddagger}$~\cite{wang2023images}} & ~ & \textcolor{gray}{29.42} & \textcolor{gray}{0.867} & \textcolor{gray}{0.164} & ~ & \textcolor{gray}{38.58} & \textcolor{gray}{0.954} & \textcolor{gray}{0.220} & ~ & \textcolor{gray}{22.34} & \textcolor{gray}{0.806} & \textcolor{gray}{0.205} \\
    Painter$^{\dagger}$~\cite{wang2023images} & ~ & \textbf{25.84} & \textbf{0.840} & \textbf{0.191} & ~ & \underline{32.84} & \textbf{0.933} & 0.224 & ~ & \underline{20.18} & \textbf{0.733} & 0.354 \\
    PromptDiffusion~\cite{wang2023context} & ~ & 21.29 & 0.568 & 0.364 & ~ & 32.33 & 0.870 & \underline{0.120} & ~ & 20.00 & 0.640 & 0.326 \\
    InstructDiffusion~\cite{geng2024instructdiffusion} & ~ & - & - & - & ~ & 29.06 & 0.820 & 0.234 & ~ & - & - & - \\
    \texttt{ICT-st} & ~ & 22.62 $\pm$ 0.02 & 0.598 $\pm$ 0.004 & 0.302 $\pm$ 0.004 & ~ & 34.55 $\pm$ 0.05 & 0.907 $\pm$ 0.007 & 0.095 $\pm$ 0.002 & ~ & \textbf{20.63} $\pm$ 0.03 & \underline{0.681} $\pm$ 0.003 & \textbf{0.256} $\pm$ 0.004 \\
    \texttt{ICT-sc} & ~ & \underline{22.64} $\pm$ 0.03 & \underline{0.599} $\pm$ 0.004 & \underline{0.301} $\pm$ 0.005 & ~ & \textbf{34.80} $\pm$ 0.04 & \underline{0.910} $\pm$ 0.005 & \textbf{0.092} $\pm$ 0.003 & ~ & 19.91 $\pm$ 0.03 & 0.665 $\pm$ 0.004 & \underline{0.286} $\pm$ 0.002 \\
    \textcolor{gray}{\texttt{ICT}$^{\uplus}$} & ~ & \textcolor{gray}{24.53} & \textcolor{gray}{0.650} & \textcolor{gray}{0.249} & ~ & \textcolor{gray}{36.56} & \textcolor{gray}{0.934} & \textcolor{gray}{0.054} & ~ & \textcolor{gray}{25.20} & \textcolor{gray}{0.729} & \textcolor{gray}{0.218} \\
    \bottomrule
  \end{tabular}
\end{table*}

\begin{table*}[!t]
\setlength{\tabcolsep}{2.6pt}
\caption{Comparison Results on Conditional Image Generation. $^{\ast}$: Specialized Methods for Each Task. $^{\dagger}$: Trained Using Official Code under the Same Computing Resources as \texttt{ICT}. Note That There Is No Officially Trained Painter Model for Conditional Image Generation. \textbf{Bold}: Best. \underline{Underline}: Second Best
}
\label{tab:comp-generation}
  \centering
    \begin{tabular}{ccccccccccccc}
    \toprule 
    \multirow{2.5}*{Method} & ~ & \multicolumn{2}{c}{Mask-to-image} & ~ & \multicolumn{2}{c}{Depth-to-image} & ~ & \multicolumn{2}{c}{Pose-to-image} & ~ & \multicolumn{2}{c}{Edge-to-image} \\
    \cmidrule{3-4}
    \cmidrule{6-7}
    \cmidrule{9-10}
    \cmidrule{12-13}
    ~ & ~ & FID$\downarrow$ & mIoU$\uparrow$ & ~ & FID$\downarrow$ & RMSE$\downarrow$ & ~ & FID$\downarrow$ & mAP$\uparrow$ & ~ & FID$\downarrow$ & SSIM$\uparrow$ \\
    \midrule
    \textcolor{gray}{ControlNet$^{\ast}$~\cite{zhang2023adding}} & ~ & \textcolor{gray}{35.4} & \textcolor{gray}{36.5} & ~ & \textcolor{gray}{43.9} & \textcolor{gray}{0.327} & ~ & \textcolor{gray}{43.0} & \textcolor{gray}{0.479} & ~ & \textcolor{gray}{12.9} & \textcolor{gray}{0.685} \\
    Painter$^{\dagger}$~\cite{wang2023images} & ~ & 75.7 & 23.2 & ~ & 89.3 & 0.417 & ~ & 200.1 & \textbf{0.709} & ~ & 233.1 & 0.271 \\
     PromptDiffusion~\cite{wang2023context}  & ~ & 31.0 & \textbf{30.5} & ~ & 52.5 & 0.528 & ~ & 40.6 & \underline{0.457} & ~ & 13.8 & \textbf{0.690} \\
    \texttt{ICT-st} & ~ & \underline{29.9} $\pm$ 0.3 & \underline{28.7} $\pm$ 0.2 & ~ & \textbf{44.0} $\pm$ 0.7 & \underline{0.374} $\pm$ 0.005 & ~ & \underline{34.7} $\pm$ 0.3 & 0.359 $\pm$ 0.009 & ~ & \underline{13.6} $\pm$ 0.2 & 0.671 $\pm$ 0.005 \\
    \texttt{ICT-sc} & ~ & \textbf{27.8} $\pm$ 0.6 & 28.5 $\pm$ 0.4 & ~ & \underline{44.2} $\pm$ 0.8 & \textbf{0.372} $\pm$ 0.004 & ~ & \textbf{34.3} $\pm$ 0.5 & 0.430 $\pm$ 0.007 & ~ & \textbf{13.5} $\pm$ 0.4 & \underline{0.673} $\pm$ 0.007 \\
    \bottomrule
  \end{tabular}
\end{table*}

\noindent \textbf{Datasets.}
We conduct experiments involving ten vision tasks on six datasets, including ADE20K~\cite{zhou2017scene}, NYUv2~\cite{silberman2012indoor}, COCO~\cite{lin2014microsoft}, merged deraining datasets~\cite{fu2017removing,yang2017deep,zhang2019image,zhang2018density,li2016rain}, SIDD~\cite{abdelhamed2018high}, and LoL~\cite{wei2018deep}. We adopt the same training/testing split as Painter~\cite{wang2023images}. Please refer to Table~\ref{tab:datasets} for a detailed task and dataset configuration.

\noindent \textbf{Methods.} We evaluate \texttt{ICT} with its two direct competitors, Painter~\cite{wang2023images} and PromptDiffusion~\cite{wang2023context}, both designed to handle multiple tasks using a unified framework, as state-of-the-art methods. Some small-scale comparisons with related works including InstructCV~\cite{DBLP:conf/iclr/GanPSPA24}, Unified-IO 2~\cite{lu2024unified}, and InstructDiffusion~\cite{geng2024instructdiffusion} are conducted as well.
We also report the results of other competing methods, which are specially trained on single tasks and do not use a general framework, for reference purposes.
Due to limited computing resources, we cannot jointly train \texttt{ICT} on data from all tasks to achieve convergence in an affordable time. Therefore, we mainly report the results of single-task models (\texttt{ICT-st}) that are separately trained for each task, and single-category models (\texttt{ICT-sc}) that are jointly trained on data from multiple tasks of the same category. Nevertheless, we train a multi-category model (\texttt{ICT-mc}) on data from three tasks belonging to distinct categories to demonstrate our \texttt{ICT}'s validity in tackling various tasks using a single set of model parameters.

\noindent \textbf{Implementation Details.}
We utilize the same training settings of SD to optimize \texttt{ICT}. We adopt a smooth L1 version of (\ref{eq2}) to train our model. We accumulate gradients every 16 batches with a batch size of 64. The learning rate is fixed to $6.4\times10^{-5}$. All training images are resized to $256\times256$ and we train \texttt{ICT} on 4 RTX 3090 GPUs. During training, we randomly drop 10\% text-conditioning to improve classifier-free guidance sampling~\cite{ho2022classifier}. We load pre-trained weights from Stable Diffusion-v1.5-inpainting for fine-tuning. We randomly sample an input-output pair as the image instruction during training and adopt a fixed pair from the training dataset (same as Painter~\cite{wang2023images}) as the image context at the inference time. 
By setting the random noise in the reverse diffusion process to 0 (\textit{i.e.}, a deterministic sampling), DDIM~\cite{song2020denoising} manages to generate an image with fewer sampling steps compared to DDPM. We adopt DDIM sampling with 50 steps for inference.

\subsection{Results and Analyses}

\begin{figure}[!t]
  \centering
    \includegraphics[width=1.0\linewidth]{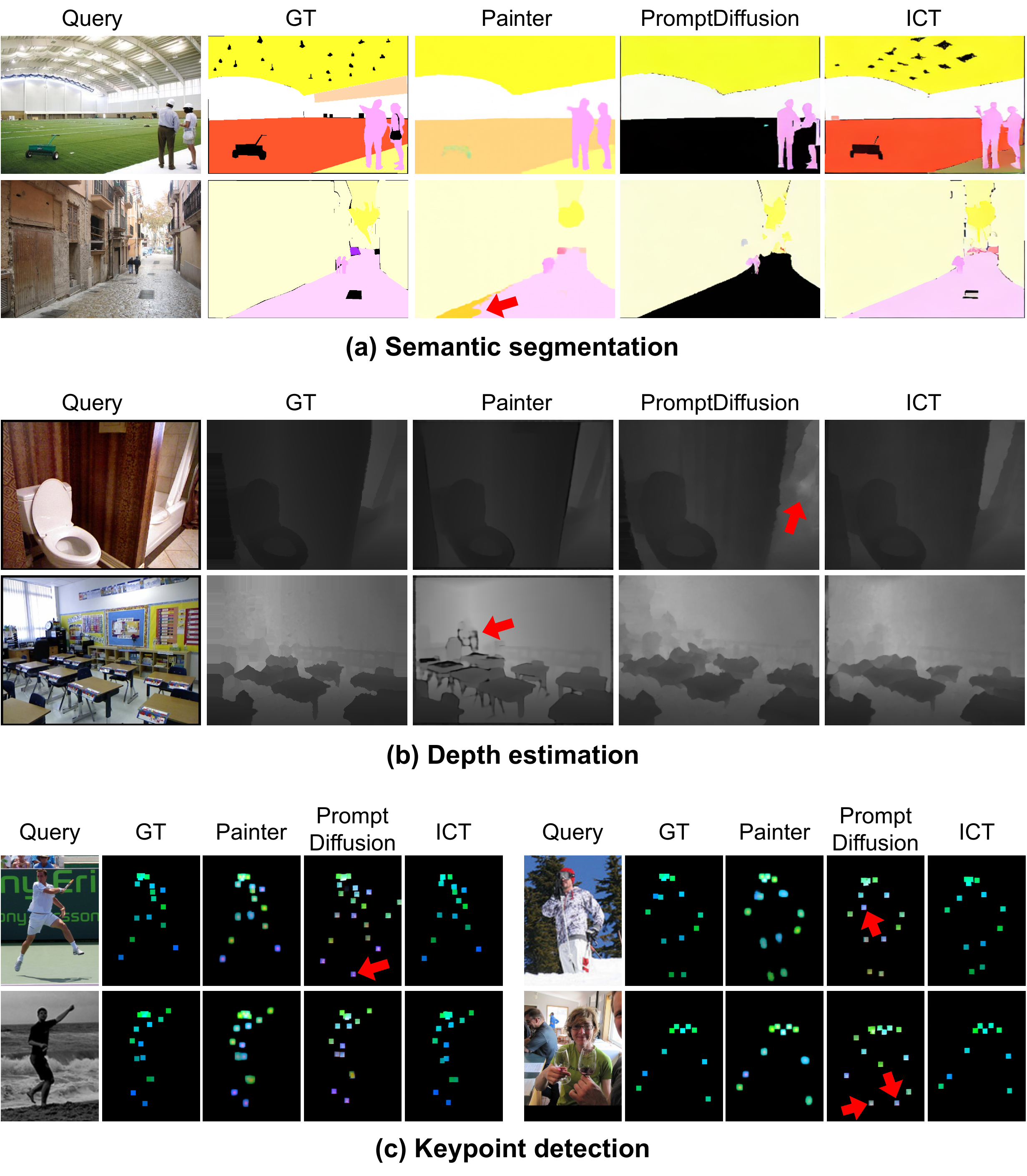}
  \caption{Visual comparison results on visual recognition. ``GT'' represents ground truth. (a) PromptDiffusion~\cite{wang2023context} tends to mispredict pixels as ``black''. Red arrow indicates the mispredictions produced by Painter~\cite{wang2023images}. (b) The competing methods, especially PromptDiffusion~\cite{wang2023context}, fail to consistently produce accurate depth predictions (red arrow). (c) PromptDiffusion~\cite{wang2023context} tends to overpredict keypoints (red arrow). In contrast, the proposed \texttt{ICT} consistently produces accurate predictions.}
  \label{fig:visual_results_recognition}
\end{figure}

\begin{figure}[!t]
  \centering
    \includegraphics[width=1.0\linewidth]{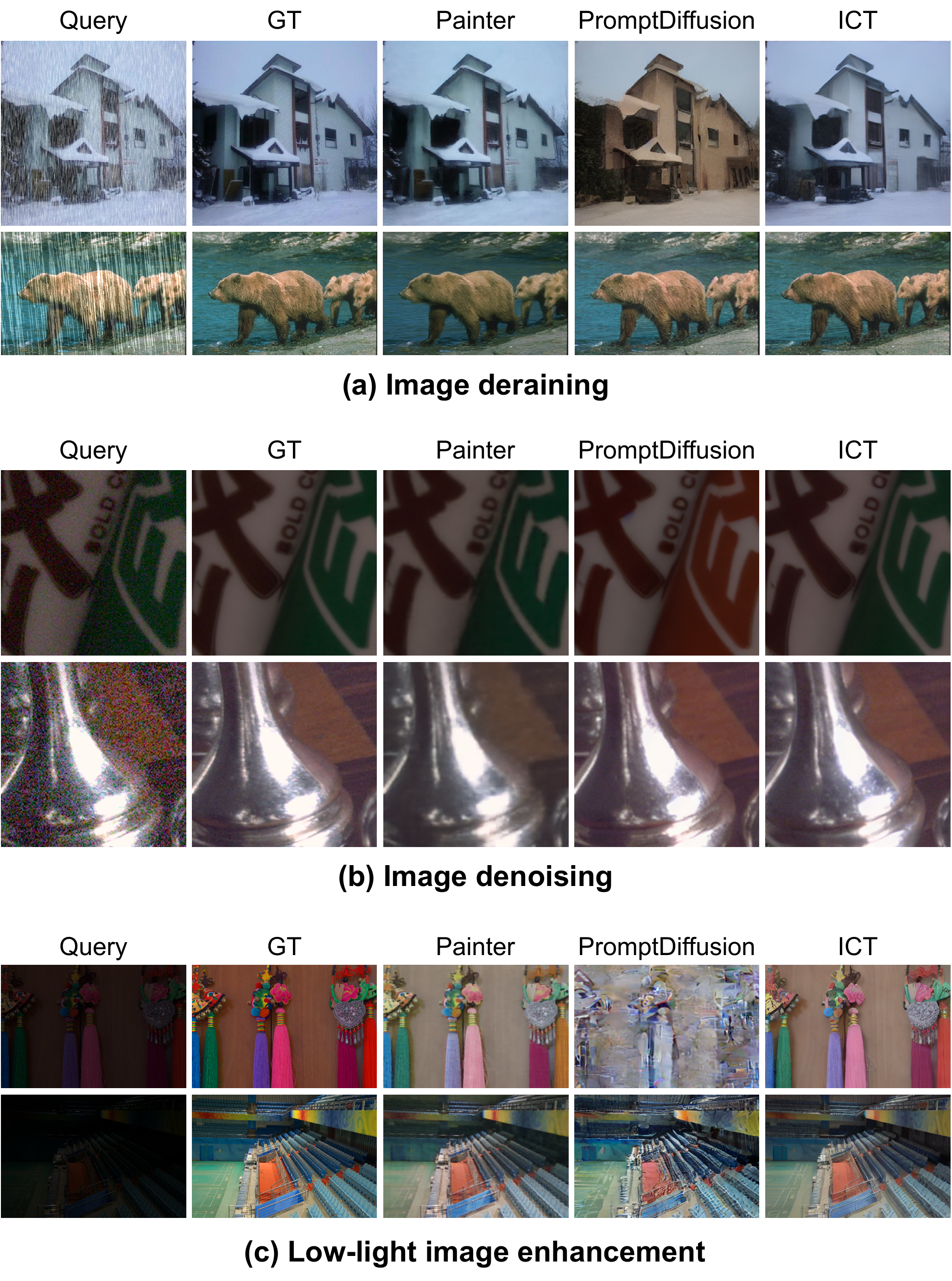}
  \caption{Visual comparison results on low-level image processing. ``GT'' represents ground truth. PromptDiffusion~\cite{wang2023context} tends to generate distorted results or images with color shifts that do not adhere to the query image.}
  \label{fig:visual_results_processing}
\end{figure}

\begin{figure}[!t]
  \centering
    \includegraphics[width=1.0\linewidth]{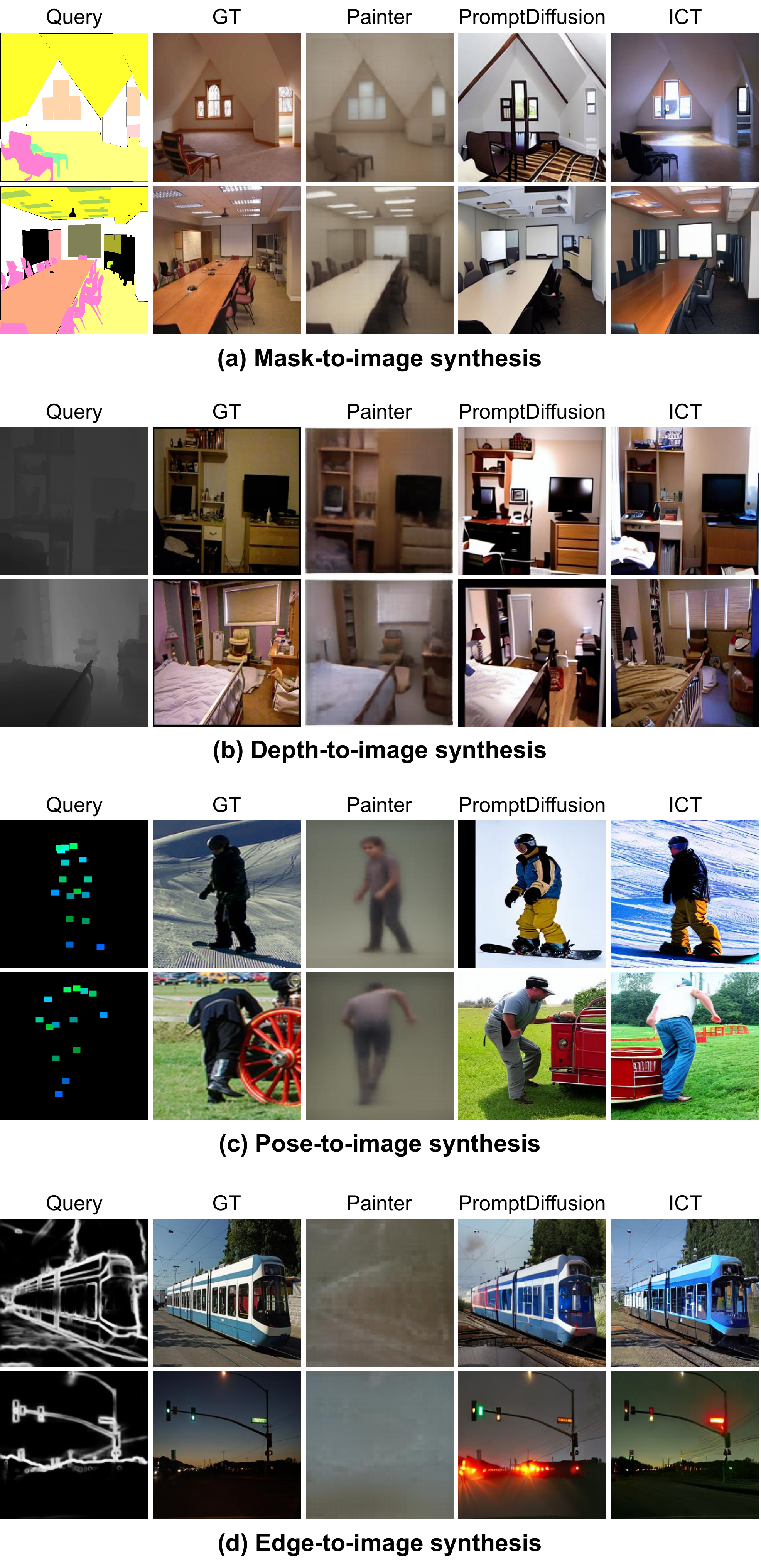}
  \caption{Visual comparison results on conditional image generation. ``GT'' represents ground truth. Painter~\cite{wang2023images} failed in generating realistic images from conditions with sparse semantics.}
  \label{fig:visual_results_generation}
\end{figure}

\noindent \textbf{Visual Recognition Results.}
We assess the proposed \texttt{ICT} on three visual recognition tasks described in Section~\ref{sec:approach_sub1}. Standard metrics are adopted for evaluation: 1)~mean Intersection-over-Union (mIoU) for semantic segmentation; 2)~root mean squared error (RMSE), absolute relative error (REL), and the accuracy under the threshold ($\delta_{1} < 1.25$) for depth estimation.

Quantitative and qualitative comparison results are presented in Table~\ref{tab:comp-prediction} and Fig.~\ref{fig:visual_results_recognition} respectively. We make the following \textbf{O}bservations.
\textbf{O1}: \texttt{ICT} outperforms PromptDiffusion by a large margin, despite both adopting the pre-trained SD as the backbone.
As shown in Fig.~\ref{fig:visual_results_recognition}(a), PromptDiffusion tends to mispredict pixels as ``black''.
We attribute \texttt{ICT}'s superior performance to our more favorable in-context image translation framework that better integrates the image context with the query image through spatial concatenation rather than the pixel-wise sum operation applied in PromptDiffusion (as we discussed in Section~\ref{sec:rel})..
\textbf{O2}: The official results of Painter are better than ours, but Painter requires training on a considerable number of machines, \textit{\textit{\textit{i.e.}}}, around 128 RTX 3090 GPUs. Hence, we retrain Painter following its official code to compare with \texttt{ICT} more fairly under the same academia-level compute (4 RTX 3090 GPUs). In this case, our \texttt{ICT} is highly comparable with Painter on visual recognition tasks. The visual results presented in Fig.~\ref{fig:visual_results_recognition} also verify that \texttt{ICT} succeeds in perceiving various scenes regarding semantics and salient object components, and subsequently produces accurate predictions in RGB format.
\textbf{O3}: \texttt{ICT} as well as Painter and PromptDiffusion still falls behind most specialized methods. However, the primary focus of this paper is to reveal the potential of generative modeling in building a general framework for three distinct categories of vision tasks,
with achieving state-of-the-art performance being of our lower priority.

It is worth noting that \texttt{ICT} performs well on keypoint detection (see Fig.~\ref{fig:visual_results_recognition}(c)). Nevertheless, generating heatmaps to calculate metrics such as average precision (AP) is difficult to accomplish with the autoencoder of pre-trained SD as it introduces lossy compression. Limited by this, we do not report quantitative results. This issue can be alleviated by resorting to better pre-trained models in the future or employing an extra model to transfer the output to heatmaps as done in InstructDiffusion~\cite{geng2024instructdiffusion}.

\noindent \textbf{Low-level Image Processing Results.}
We exploit the ability of \texttt{ICT} to perform low-level image processing on three image restoration tasks. Standard metrics, including PSNR, SSIM, and LPIPS~\cite{zhang2018unreasonable}, are used for evaluation.

Table~\ref{tab:comp-lowlevel} presents the quantitative comparison results. Similar to the observations in visual recognition, here \texttt{ICT} attains competitive performance compared to Painter (retrained version) and surpasses PromptDiffusion in all metrics.
It is worth noting that there is an upper bound for \texttt{ICT} because the autoencoder of pre-trained SD brings information loss (as pointed out in InstructDiffusion~\cite{geng2024instructdiffusion}). We apply the autoencoder to reconstruct the ground truth and calculate the metrics as our upper bound.
In addition, \texttt{ICT-sc} produces better results than \texttt{ICT-st} on image deraining and denoising, indicating the capability of \texttt{ICT-sc} to perform versatile low-level image processing using a single model. 
Visual results illustrated in Fig.~\ref{fig:visual_results_processing} also demonstrate the efficacy of \texttt{ICT} in handling low-level image processing tasks. 

\noindent \textbf{Conditional Image Generation Results.}
We evaluate the conditional image generation performance of \texttt{ICT} given various conditions, including segmentation mask, depth map, keypoint, and HED edge. The commonly used Fr\'echet Inception Distance (FID)~\cite{heusel2017gans} is adopted to assess the realism of the generated images. To evaluate the alignment of the generated images with the given conditions, we follow Uni-ControlNet~\cite{zhao2024uni} to employ the following metrics for four different conditional image generation tasks: 1)~mean Intersection-over-Union (mIoU) for mask-to-image synthesis; 2)~root mean squared error (RMSE) for depth-to-image synthesis; 3)~mean Average Precision (mAP) for pose-to-image synthesis; 4)~SSIM for edge-to-image synthesis. We calculate these metrics by comparing the extracted conditions from the generated images and the given conditions.

The comparison results are presented in Table~\ref{tab:comp-generation} and Fig.~\ref{fig:visual_results_generation}. \textbf{O}bservations are elaborated in the following.
\textbf{O1}: The proposed \texttt{ICT} achieves exceptional performance on all tasks and even surpasses the specialized method (ControlNet) on mask/depth/pose-to-image synthesis, indicating that \texttt{ICT} fully unleashes the generative power of pre-trained SD.
\textbf{O2}: Painter, which is built on top of pre-trained MAE, falls short of synthesizing realistic images from conditions with sparse semantics, resulting in poor FID values. The blurry images synthesized by Painter shown in Fig.~\ref{fig:visual_results_generation} further validate the collapse of Painter on conditional image generation tasks.
\textbf{O3}: \texttt{ICT-sc} attains a higher performance than \texttt{ICT-st} on most of conditional image generation tasks. This showcases the effectiveness of \texttt{ICT-sc} in translating flexible control signals into high-fidelity images using a unified model.
As presented in Fig.~\ref{fig:visual_results_generation}, given different conditions, \texttt{ICT-sc} manages to recognize the underlying task and synthesize photorealistic images that spatially conform to the control signals.

\begin{table*}[!t]
\setlength{\tabcolsep}{3.92pt}
\caption{Results of Our Multi-Category Model \texttt{ICT-mc}. We Select Three Representative Tasks from Three Categories to Investigate \texttt{ICT-mc}}
\label{tab:joint-train-results}
  \centering
    \begin{tabular}{cccccccccccc}
    \toprule
    \multirow{2.5}*{Method} & ~ & \multicolumn{3}{c}{Depth estimation} & ~ & \multicolumn{3}{c}{Denoising} & ~ & \multicolumn{2}{c}{Mask-to-image} \\
    \cmidrule{3-5}
    \cmidrule{7-9}
    \cmidrule{11-12}
    ~ & ~ & RMSE$\downarrow$ & REL$\downarrow$ & $\delta_{1}$$\uparrow$ & ~ & PSNR$\uparrow$ & SSIM$\uparrow$ & LPIPS$\downarrow$ & ~ & FID$\downarrow$ & mIoU$\uparrow$ \\
    \midrule
    \texttt{ICT-st} & ~ & 0.420 & 0.135 & 0.857 & ~ & 34.55 & 0.907 & 0.095 & ~ & 29.9 & 28.7 \\
    \texttt{ICT-mc} & ~ & 0.421 & 0.131 & 0.863 & ~ & 34.58 & 0.909 & 0.095 & ~ & 30.4 & 33.2 \\
    \bottomrule
  \end{tabular}
\end{table*}

\noindent \textbf{Results of Multi-category Model.}
We collect data from depth estimation, denoising, and mask-to-image synthesis to jointly train a multi-category model \texttt{ICT-mc}. As shown in Table~\ref{tab:joint-train-results}, \texttt{ICT-mc} achieves a competitive performance very close to \texttt{ICT-st}, confirming the proposed framework's ability to automatically identify the specific task through the given context and produce the corresponding desired output.
It is encouraging to see the results of \texttt{ICT-mc} trained for these tasks involving disparate visual signals and data domains, and we believe that unifying discrimination and generation via a single model will be made possible if the proposed \texttt{ICT} can be trained with sufficient computational resources.

\begin{table*}[!t]
\setlength{\tabcolsep}{2.0pt}
\caption{Ablation Study Results of \texttt{ICT-st} for Semantic Segmentation on ADE20K}
\label{tab:ablation}
  \centering
    \begin{tabular}{ccccccccccc}
    \toprule  
    \multirow{2.5}*{Method} & ~ & \multicolumn{2}{c}{Masking strategy} & ~ & \multicolumn{3}{c}{Type of text context} & ~ & \multicolumn{2}{c}{Type of image context} \\
    \cmidrule{3-4}
    \cmidrule{6-8}
    \cmidrule{10-11}
    ~ & ~ & region-wise & whole image & ~ & no prompt & task-level & instance-level & ~ & fixed pair & random pair \\
    \midrule
    mIoU$\uparrow$ & ~ & 17.4 & \textbf{33.4} & ~ & 31.0 & 31.1 & \textbf{33.4} & ~ & 33.4 & \textbf{33.7} \\
    \bottomrule
  \end{tabular}
\end{table*}

\begin{table*}[!t]
\setlength{\tabcolsep}{2.6pt}
\caption{Ablation Study Results of \texttt{ICT-mc} Regarding Text Context}
\label{tab:text_context_ablation}
  \centering
    \begin{tabular}{cccccccccccc}
    \toprule
    \multirow{2.5}*{Method} & ~ & \multicolumn{3}{c}{Depth estimation} & ~ & \multicolumn{3}{c}{Denoising} & ~ & \multicolumn{2}{c}{Mask-to-image} \\
    \cmidrule{3-5}
    \cmidrule{7-9}
    \cmidrule{11-12}
    ~ & ~ & RMSE$\downarrow$ & REL$\downarrow$ & $\delta_{1}$$\uparrow$ & ~ & PSNR$\uparrow$ & SSIM$\uparrow$ & LPIPS$\downarrow$ & ~ & FID$\downarrow$ & mIoU$\uparrow$ \\
    \midrule
    \texttt{ICT-mc} w/ text context & ~ & 0.421 & 0.131 & 0.863 & ~ & 34.58 & 0.909 & 0.095 & ~ & 30.4 & 33.2 \\
    \texttt{ICT-mc} w/o text context & ~ & 0.466 & 0.154 & 0.826 & ~ & 34.36 & 0.907 & 0.101 & ~ & 31.5 & 29.5 \\
    \bottomrule
  \end{tabular}
\end{table*}

%
\subsection{Ablation Study}
We perform ablations on three ingredients of \texttt{ICT}: the masking strategy, the design of text context, and the type of image context.

\noindent \textbf{Masking Strategy.}
Instead of masking the whole image $I_{gt}$ during training, we randomly mask out a portion of $I_{gt}$ to train \texttt{ICT} for the semantic segmentation task. As reported in Table~\ref{tab:ablation}, this region-wise masking results in a significant performance drop, highlighting the importance of our masking strategy in unleashing the unifying ability of pre-trained SD.

\noindent \textbf{Design of Text Context.}
We also study the utility of text context by conducting two ablations for \texttt{ICT-st} and \texttt{ICT-mc} respectively. Results are provided in Table~\ref{tab:ablation} and Table~\ref{tab:text_context_ablation}. As can be seen, text context is beneficial for some tasks (\textit{e.g.}, depth estimation), but the gain by applying text context is very marginal for other tasks (\textit{e.g.}, denoising). Therefore, one can optionally apply text context to strike a balance between high-end performance and extra human efforts.

\noindent \textbf{Type of Image Context.}
We further investigate the effect of image context by comparing two methods of selecting image context during inference: 1)~randomly selecting an input-output pair from the training data each time, and 2)~adopting a fixed input-output pair from the training data for all testing samples. As shown in Table~\ref{tab:ablation}, applying random selection of image context attains better performance than using fixed context. This implies the potential of improving the performance of our work by finding better image context.

\begin{figure}[!t]
  \centering
    \includegraphics[width=1.0\linewidth]{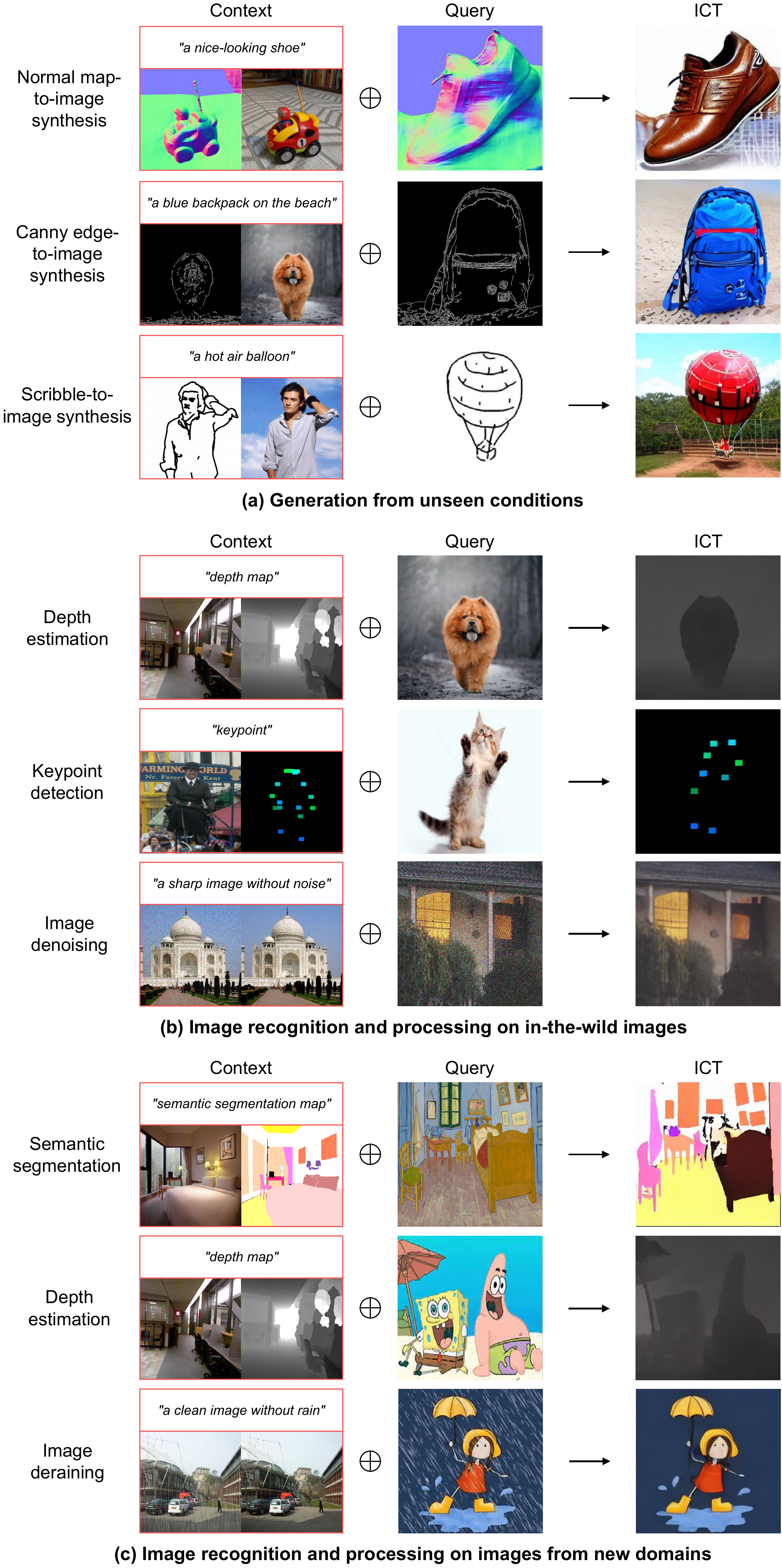}
  \caption{Example results from our \texttt{ICT} in performing out-of-distribution inference.}
  \label{fig:generalization_results}
\end{figure}

\begin{figure}[!t]
  \centering
    \includegraphics[width=1.0\linewidth]{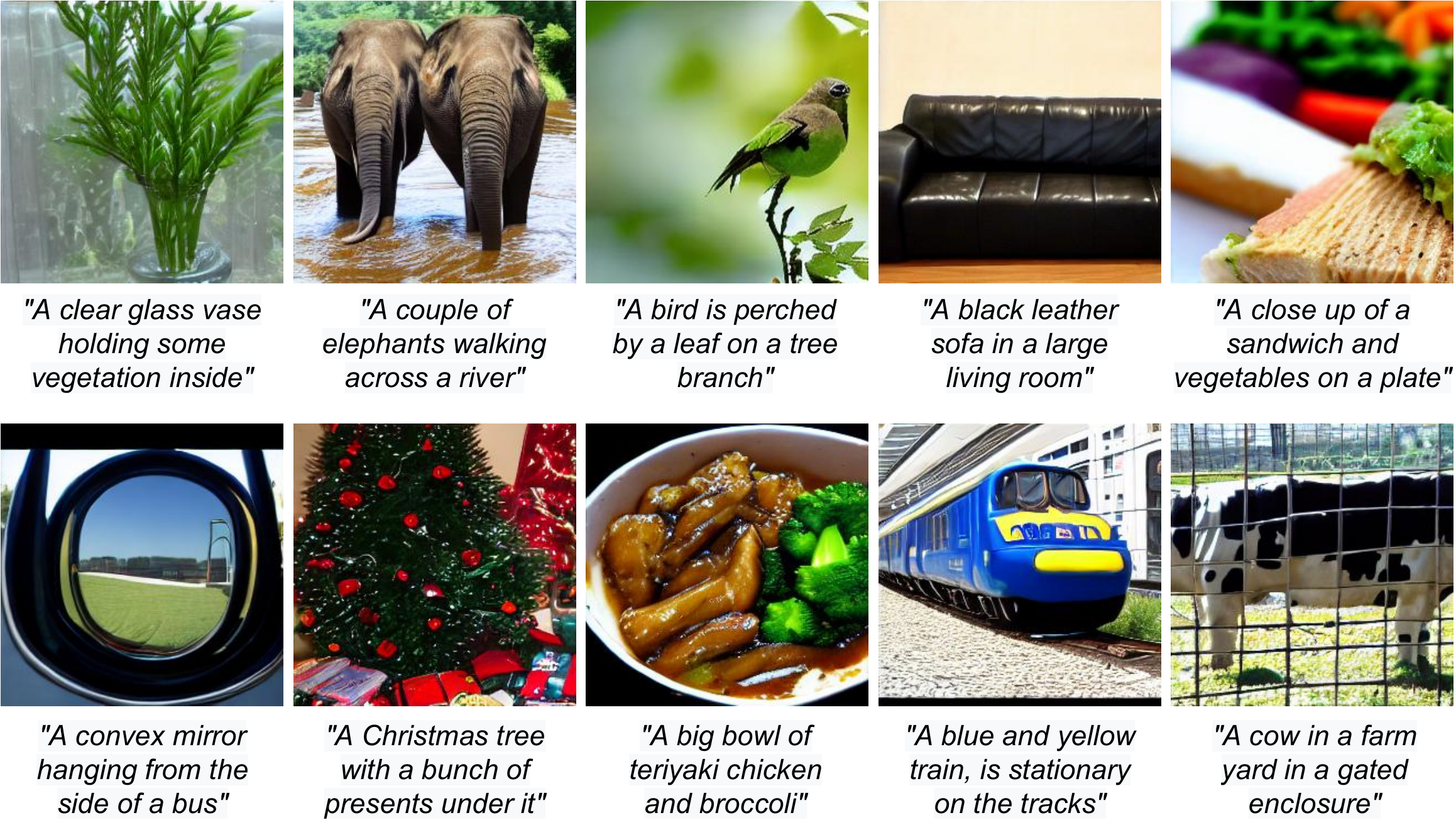}
  \caption{Text-to-image generation results of \texttt{ICT}.}
  \label{fig:addi_visual_results_T2I}
\end{figure}

\begin{table}[!t]
\setlength{\tabcolsep}{5.0pt}
\caption{Comparison Results on Out-of-Distribution Datasets}
\label{tab:generalization}
  \centering
    \begin{tabular}{ccccc}
    \toprule  
    \multirow{2.5}*{Method} & ~ & \multicolumn{3}{c}{\makecell[c]{Depth estimation \\ on SUN RGB-D~\cite{song2015sun}}} \\
    \cmidrule{3-5}
    ~ & ~ & RMSE$\downarrow$ & REL$\downarrow$ & $\delta_{1}$$\uparrow$ \\
    \midrule
    Painter~\cite{wang2023images} & ~ & \underline{0.515} & \textbf{0.272} & \underline{0.766} \\
    PromptDiffusion~\cite{wang2023context} & ~ & 0.847 & 0.357 & 0.652 \\
    \texttt{ICT} & ~ & \textbf{0.479} & \underline{0.293} & \textbf{0.768} \\
    \bottomrule
  \end{tabular}
\end{table}

\begin{table}[!t]
\setlength{\tabcolsep}{3pt}
\caption{Results of \texttt{ICT} Using Different Shots of Image Context}
\label{tab:few-shot}
  \centering
    \begin{tabular}{ccccccccc}
    \toprule 
    \multirow{2.5}*{\makecell[c]{\# of shots \\ during inference}} & ~ & \multicolumn{3}{c}{Depth estimation} & ~ & \multicolumn{3}{c}{Denoising}\\
    \cmidrule{3-5}
    \cmidrule{7-9}
    ~ & ~ & RMSE$\downarrow$ & REL$\downarrow$ & $\delta_{1}$$\uparrow$ & ~ & PSNR$\uparrow$ & SSIM$\uparrow$ & LPIPS$\downarrow$ \\
    \midrule
    one-shot & ~ & 0.420 & 0.135 & 0.857 & ~ & 34.55 & 0.907 & 0.095 \\
    two-shot & ~ & 0.416 & 0.131 & 0.863 & ~ & 34.73 & 0.911 & 0.095 \\
    four-shot & ~ & 0.413 & 0.130 & 0.863 & ~ & 34.75 & 0.911 & 0.095 \\
    \bottomrule
  \end{tabular}
\end{table}

\begin{table}[!t]
\setlength{\tabcolsep}{2.2pt}
\caption{Scaling Behavior of \texttt{ICT} w.r.t. Computing Resources}
\label{tab:scaling}
  \centering
    \begin{tabular}{ccccccccc}
    \toprule 
    \multirow{2.5}*{Computing resources} & ~ & \multicolumn{3}{c}{Depth estimation} & ~ & \multicolumn{3}{c}{Denoising} \\
    \cmidrule{3-5}
    \cmidrule{7-9}
    ~ & ~ & RMSE$\downarrow$ & REL$\downarrow$ & $\delta_{1}$$\uparrow$ & ~ & PSNR$\uparrow$ & SSIM$\uparrow$ & LPIPS$\downarrow$ \\
    \midrule
    one 3090 GPU & ~ & 0.461 & 0.156 & 0.812 & ~ & 34.02 & 0.898 & 0.121 \\
    four 3090 GPUs & ~ & 0.420 & 0.135 & 0.857 & ~ & 34.55 & 0.907 & 0.095 \\
    eight A100 GPUs & ~ & 0.391 & 0.118 & 0.892 & ~ & 34.92 & 0.913 & 0.092 \\
    \bottomrule
  \end{tabular}
\end{table}

\subsection{Generalization Capability}
We explore \texttt{ICT}'s generalization capability by applying it to unseen tasks/data. As demonstrated in Fig.~\ref{fig:generalization_results}, \texttt{ICT} is capable of 1)~generating realistic images from conditions that are unseen during training (\textit{e.g.}, normal map, canny edge, and scribble), 2)~conducting image recognition and processing (\textit{e.g.}, depth estimation, keypoint detection, and image denoising) on in-the-wild images that have different data distribution compared to the training dataset, and 3)~performing image recognition and processing (\textit{e.g.}, semantic segmentation, depth estimation, and image deraining) on images from new domains (\textit{e.g.}, Van Gogh's paintings in Fig.~\ref{fig:generalization_results}(c)).
These promising results indicate that \texttt{ICT} learns the underlying ``structure'' of various visual signals and generalizes well to new scenarios by leveraging the pre-trained knowledge to conduct in-context image translation.

In addition, we evaluate different methods on unseen datasets and report the quantitative results in Table~\ref{tab:generalization}. Compared to other competing methods, \texttt{ICT} exhibits superior performance on data unseen in training, indicating the strong generalization ability of our proposed method.

We also explore an additional task: text-conditional image generation. This task can be fulfilled by directly applying \texttt{ICT-sc} trained on four conditional image generation tasks where the query is set to a black image. We achieve a FID of 27.47 on the COCO validation set. This could be further improved by training a \texttt{ICT-st} on this task. We present some text-to-image generation results obtained from our method in Fig.~\ref{fig:addi_visual_results_T2I}.

\subsection{Few-shot In-context Inference}
Here we extend \texttt{ICT} to perform few-shot in-context inference. This is fulfilled by establishing the grid image with more input-output pairs. We report the results of \texttt{ICT} under one-shot, two-shot, and four-shot settings in Table~\ref{tab:few-shot}. We observe that there is a slight gain by introducing more image context, and we think this could be further explored during training, which we leave to future work.

\subsection{Efficiency and Scalability}
Under limited compute (\textit{i.e.}, 4 RTX 3090 GPUs), \texttt{ICT} exhibits better performance (especially its unification of more categories of tasks) compared to its competitors, which showcases the higher efficiency of the proposed framework.
We further conduct an experiment where \texttt{ICT} is trained using different amounts of computing resources. We present the results in Table~\ref{tab:scaling}. \texttt{ICT} shows impressive scaling behavior where the performance improves with increasing computing power.

\section{Conclusion}
\label{sec:con}

In this paper, we explore the trajectory of LLMs to design a general vision framework for visual recognition, low-level image processing, and conditional image generation, identifying two essential components: 1)~a unified data format, and 2)~a general training pipeline.
To this end, the proposed \texttt{ICT} achieves the unification by 1)~framing heterogeneous visual signals as RGB images, and 2)~introducing a new in-context image translation pipeline to standardize the training of various tasks.
Compared to the evaluated competitors that collapse on at least one category of vision tasks, \texttt{ICT} attains competitive performance across three categories including ten vision tasks in total. We hope these results could encourage further study on how to leverage generative modeling to perceive and process visual signals in a general manner.

 

\bibliographystyle{IEEEtran}
\bibliography{main}


 




\vfill

\end{document}